\documentclass[letterpaper]{article} 
\usepackage{aaai25}  
\usepackage{times}  
\usepackage{helvet}  
\usepackage{courier}  
\usepackage[hyphens]{url}  
\usepackage{graphicx} 
\usepackage{natbib}  
\usepackage{caption} 
\usepackage{algorithm}
\usepackage{algorithmic}
\usepackage{amsmath}
\usepackage{amsthm}
\usepackage{amssymb}
\usepackage{mathtools}
\usepackage{booktabs}
\usepackage{boldline} 
\usepackage{multirow}
\usepackage{colortbl}
\usepackage{xcolor}
\usepackage{graphicx}
\usepackage{newfloat}
\usepackage{listings}
\DeclareCaptionStyle{ruled}{labelfont=normalfont,labelsep=colon,strut=off} 
\floatstyle{ruled}
\newfloat{listing}{tb}{lst}{}
\floatname{listing}{Listing}
\title{Virtual Nodes Can Help: Tackling Distribution Shifts \\ in Federated Graph Learning}
\author{
    Xingbo Fu, Zihan Chen, Yinhan He, Song Wang, Binchi Zhang, Chen Chen, Jundong Li 
}
\affiliations{

    University of Virginia\\
    \{xf3av, brf3rx, nee7ne, sw3wv, epb6gw, zrh6du, jundong\}@virginia.edu
}

\usepackage{bibentry}

\begin{document}

\maketitle

\begin{abstract}
Federated Graph Learning (FGL) enables multiple clients to jointly train powerful graph learning models, e.g., Graph Neural Networks (GNNs), without sharing their local graph data for graph-related downstream tasks, such as graph property prediction. In the real world, however, the graph data can suffer from significant distribution shifts across clients as the clients may collect their graph data for different purposes. In particular, graph properties are usually associated with invariant label-relevant substructures (i.e., subgraphs) across clients, while label-irrelevant substructures can appear in a client-specific manner. The issue of distribution shifts of graph data hinders the efficiency of GNN training and leads to serious performance degradation in FGL. 
To tackle the aforementioned issue, we propose a novel FGL framework entitled FedVN that eliminates distribution shifts through client-specific graph augmentation strategies with multiple learnable Virtual Nodes (VNs). Specifically, FedVN lets the clients jointly learn a set of shared VNs while training a global GNN model. To eliminate distribution shifts, each client trains a personalized edge generator that determines how the VNs connect local graphs in a client-specific manner.
Furthermore, we provide theoretical analyses indicating that FedVN can eliminate distribution shifts of graph data across clients. Comprehensive experiments on four datasets under five settings demonstrate the superiority of our proposed FedVN over nine baselines.
\end{abstract}
\begin{links}
    \link{Code}{https://github.com/xbfu/FedVN}
\end{links}

\section{Introduction}
Graphs are pervasive in a wide range of real-world scenarios, including bioinformatics \cite{yang2020graph, yuan2020gcng}, healthcare systems \cite{cui2020deterrent, fu2023spatial}, and fraud detection \cite{motie2024financial, zheng2024survey}. 
As a dominant approach for modeling graph data, Graph Neural Networks (GNNs) \cite{hamilton2017graphsage, kipf2016gcn, velivckovic2017gat, xu2018gin} have demonstrated great prowess in graph representation learning and have been widely adopted in various graph-level applications, such as molecular property regression \cite{li2022geomgcl, zhuang2024imold} and 3D shape classification \cite{wang2019dynamic, zhou2021adaptconv}.
Specifically, GNNs follow a message-passing mechanism that recursively aggregates neighboring information of each node to obtain its node embedding. The representation of an entire graph can be obtained through different graph pooling operations \cite{xu2018gin, ying2018hierarchical}, e.g., by summing all node embeddings in the graph. The final graph representation can then be used for graph property prediction, including graph classification and graph regression.
Despite their superior performance, most GNNs are trained in a centralized manner where graph data need to be collected in a single machine before training. 
In the real world, however, a great number of graph data are stored by different data owners and cannot be shared due to data privacy \cite{voigt2017gdpr,wang2024safety}, which hampers us in training powerful GNNs.

Federated Learning (FL) \cite{mcmahan2017fl} is a decentralized learning paradigm where multiple clients collaboratively train machine learning models over their local data while preserving privacy. 
In this study, we focus on Federated Graph Learning (FGL) \cite{fu2022fgml} that aims to train GNNs over distributed graph data from multiple clients in a federated fashion.
One critical challenge in the context of FGL is that graph data can encounter significant distribution shifts across clients. In graph property prediction, the property of a graph (e.g., a molecule) is usually determined by its causal substructure (e.g., functional groups) \cite{lucic2022cf, luo2020parameterized, yuan2021explainability}. Typically, such causality can be consistent across all clients. Nonetheless, the non-causal substructure in a graph can vary significantly across clients in that the clients may collect graph data for different purposes. 
Consider a medical system with two institutes as an example. Institute A studies benzene-based compounds so it mainly has molecules containing the phenyl group (e.g., benzoic acid molecules). In contrast, Institute B studies ester-based compounds so it mainly has molecules containing an ester (e.g., glyceride molecules). The goal of the two institutes is to jointly train a GNN model to predict the water solubility of molecules. Typically, the label (i.e., the water solubility) of each molecule is causally determined by the substructure hydroxy (i.e., -OH), which is invariant across the two institutes, while the function groups - the phenyl group in Institute A and the ester in Institute B - are non-causal substructures and quite different across the two clients. Such distribution shifts of graph data caused by distinct non-causal substructures incur divergent information embedded in graph representations, which results in severe performance degradation of GNNs and consequently hinders further FGL deployments in practice.

Although numerous recent efforts attempt to grapple with the distribution shift of graph data from multiple sources for graph property prediction \cite{chen2022learning, gui2024joint, liu2022graph, sui2024unleashing, zhuang2024imold}, these approaches are inapplicable to FGL since they require the multi-source graph data collected centrally in a mini-batch when training GNNs. In the context of FGL, only a few studies investigate the problem of graph property prediction in FGL \cite{tan2023fedstar, xie2021gcfl}. These studies propose to leverage common graph properties from different datasets or even divergent domains and enhance collaborative training of GNNs by either client clustering \cite{xie2021gcfl} or sharing structural knowledge \cite{tan2023fedstar}. Nevertheless, in-depth analyses of FGL with the distribution shift of graph data have not been fully explored yet.

Our goal in this study is to essentially grapple with the aforementioned challenge at the data level. More concretely, each client will learn to manipulate its original graph data so that the altered graphs can facilitate collaborative training of GNNs. Hence, we naturally ask a question: \textit{How to properly augment local graphs in each client so that the distribution shift of graph data in FGL can be eliminated?}
To answer this question, we first need to devise a suitable graph augmentation strategy for this setting. As a common trick for graph augmentation, adding a Virtual Node (VN) to graphs has been adopted by recent studies \cite{hu2021ogb, hu2020open} for graph property prediction. Typically, the VN is designed to uniformly connect to all original graph nodes and consequently improve the capacity of GNNs for capturing long-range dependencies in a graph \cite{cai2023connection}.
While the problem investigated in this paper is essentially different from the above studies, the role of VNs in FGL remains an untouched area that is worthy of extensive exploration.

In this study, we propose a novel FGL framework entitled FedVN to eliminate distribution shifts of graph data across clients. The intuition of FedVN is to let each client learn a client-specific graph augmentation strategy and enable global GNNs to be trained over identical augmented graphs across clients. Inspired by the idea of adding VNs, we propose to introduce multiple learnable VNs that are shared across clients in FedVN. To eliminate cross-client distribution shifts of graph data, each client augments its local graphs with these VNs by a personalized edge generator. To avoid collapsing to fewer VNs, we design a decoupling loss in FedVN that encourages VNs to diffuse across the feature space. Additionally, we propose a novel score-contrastive loss to guide the generated edges within a client following the same pattern. We provide theoretical analyses to show that our design in FedVN effectually eliminates the distribution shift issue of graph data in FGL. We conduct extensive empirical evaluations on four datasets under five settings. The results demonstrate that our proposed FedVN outshines nine SOTA baselines.

Our main contributions are summarized as follows:
\begin{itemize}
    \item We take the first step towards investigating the distribution shift issue of graph data across clients for graph property prediction in FGL. In this work, we present a formal problem formulation of the studied issue.
    \item We propose a novel FGL framework FedVN to tackle the above problem by learning client-specific graph augmentation strategies with multiple VNs. We design a personalized edge generator that determines how the VNs connect local graphs so that the global GNN model can be trained over identical augmented graphs across clients.
    \item We provide theoretical analyses to support that our design in FedVN has the capability of eliminating the distribution shift issue in FGL.
    \item Extensive experiments are conducted on four datasets under five settings. The results validate the superiority of our proposed FedVN compared with other baselines.
\end{itemize}

\section{Related Work}
\subsubsection{Federated graph learning.} 
A plethora of recent efforts attempt to apply FL techniques to graph data and train powerful GNNs for various downstream tasks \cite{fu2024fedgls, fu2024fedspray}. For example, FedLit \cite{xie2023fedlit} investigates the issue of latent link-type heterogeneity for node classification. A few studies propose to recover cross-client missing information for training GNNs \cite{peng2022fedni, zhang2021fedsage}. In terms of graph property prediction, the pioneering work \cite{xie2021gcfl} demonstrates that different graphs from different datasets or even different domains may share common properties. A subsequent work \cite{tan2023fedstar} proposes to design GNNs in a feature-structure decoupled manner and share the structure encoder across clients. However, the above two works do not perform in-depth analyses of the distribution shift issue with its impact on graph property prediction in FGL. One recent work \cite{wan2024federated} investigates FGL under domain shifts but only focuses on node-level tasks.

\begin{figure*}[ht]
\centering
\includegraphics[width=\linewidth]{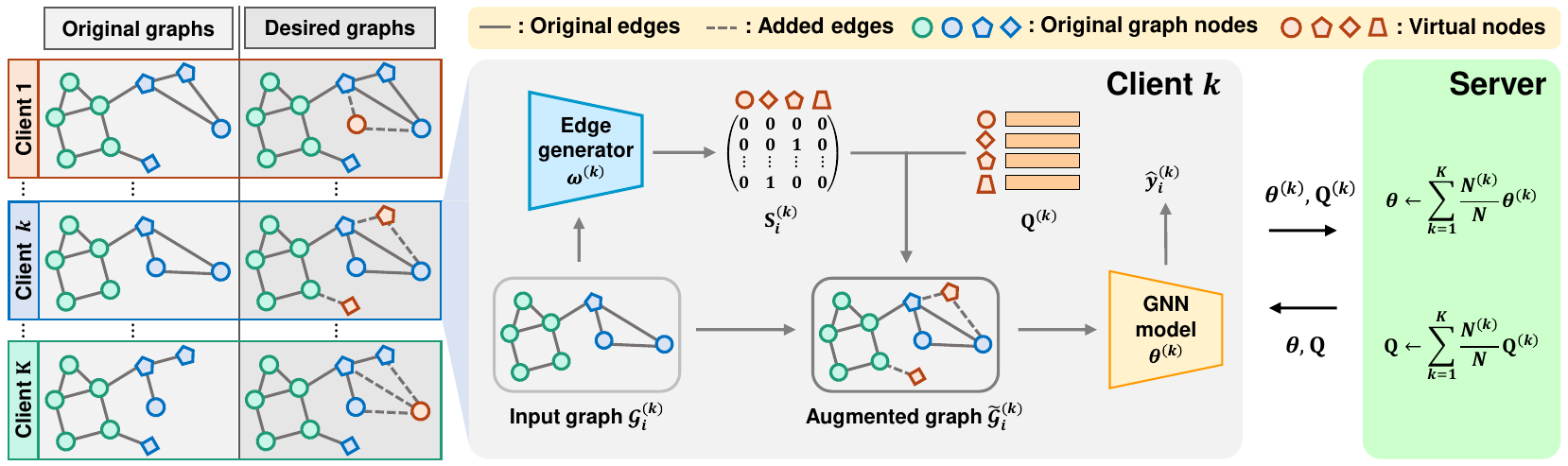}
\caption{An overview of the proposed FedVN. FedVN aims to learn client-specific graph augmentation strategies by adding multiple virtual nodes through personalized edge generators so that the global GNN model can be trained over identical graphs.}
\label{fig:framework}
\end{figure*}

\subsubsection{Virtual nodes in GNNs.}
A VN augments a graph by adding an extra node that is uniformly connected to all nodes in the original graphs \cite{gilmer2017neural}. It is demonstrated that adding the VN is effective for graph property prediction in a handful of benchmarks \cite{hu2021ogb, hu2020open} since it can help GNNs capture long-range dependencies in a graph \cite{cai2023connection}. Additionally, a recent study \cite{hwang2022analysis} investigates the role of VNs in link prediction and proposes to add multiple VNs based on node clusters.

\subsubsection{Contrastive learning.} 
As a representative scheme in self-supervised learning, contrastive learning has been widely adopted in a variety of domains, such as computer vision \cite{chen2020simple, zbontar2021barlow} and graph mining \cite{xu2024learning, shen2023neighbor, xu2023cldg}. The key idea of contrastive learning is to maximize the agreement between samples with shared semantic information (i.e., positive pairs) and minimize the agreement between irrelevant samples (i.e., negative pairs). In the context of FL, a few studies propose contrastive learning to enhance collaborative training based on different designs of sample pairs. For instance, FedPCL \cite{tan2022federated} performs contrastive learning between prototypes from different clients. FGSSL \cite{huang2023federated} designs federated node semantic contrast from global and local views.

\section{Preliminaries}
\subsection{Graph Neural Networks}
Let $\mathcal{G}=(\mathcal{V}, \mathcal{E}, \textbf{X})$ denote a graph where $\mathcal{V}$ is the set of $|\mathcal{V}|$ nodes, $\mathcal{E}$ is the edge set, and $\textbf{X}\in \mathbb{R}^{|\mathcal{V}|\times d_x}$ is the node feature matrix. $d_x$ is the number of node features. In this study, we consider graph property prediction, i.e., predicting the graph-specific target $y$ of a graph $\mathcal{G}$.
Generally, GNNs follow a message-passing mechanism \cite{hamilton2017graphsage, kipf2016gcn, velivckovic2017gat, xu2018gin} where each node in a graph iteratively aggregates information from its neighboring graph nodes to update its node representation. More specifically, the $l$-th layer of an $L$-layer GNN updates the representation of node $v \in \mathcal{V}$ by
\begin{equation} \label{gnn}
    \textbf{h}^{(l)}_v = \textrm{CB}_\text{gn}^{(l)}(\textbf{h}_v^{(l-1)}, \textrm{AGG}_{\text{gn}}^{(l)}(\{\textbf{h}_u^{(l-1)}: u \in \mathcal{N}(v)\})),
\end{equation}
where $\textrm{AGG}_{\text{gn}}^{(l)}(\cdot)$ represents the aggregation operation extracting the neighboring information of node $v$, $\textrm{CB}_\text{gn}^{(l)}(\cdot)$ represents the combination operation integrating the previous representation of node $v$ and its neighboring information, and $\mathcal{N}(v)$ denotes the neighbors of node $v$. $\textbf{h}^{(0)}_v$ is initialized with node $v$'s feature $\textbf{x}_v \in \textbf{X}$.
After the $L$-layer propagation, the \textrm{READOUT} operation pools the final node representations in a graph $\mathcal{G}$ through a permutation-invariant pooling function (e.g., $sum$ or $mean$) to obtain its representation $\textbf{h}_\mathcal{G}$, and its prediction $\hat{y}$ is computed based on its representation through a multi-layer perceptron (MLP) by
\begin{equation}
    \hat{y} = \textrm{MLP}(\textbf{h}_\mathcal{G}), \; \textrm{where} \; \textbf{h}_\mathcal{G} = \textrm{READOUT} (\{ \textbf{h}_v^{(L)}: v \in \mathcal{V}\}).
\end{equation}
In this study, we use a GNN model for graph property prediction. The GNN model typically consists of a GNN (e.g., GIN \cite{xu2018gin}) as the encoder and an MLP as the prediction head, with the READOUT operation between them.

\subsection{Federated Learning}
We consider an FL system with a set of $K$ clients. Each client $k$ has its local dataset $\mathcal{D}^{(k)}=\{(x_i^{(k)}, y_i^{(k)})\}_{i=1}^{N^{(k)}}$ where $x_i^{(k)}$ is the input and $y_i^{(k)}$ is its corresponding ground-truth label. $N^{(k)}$ is the number of data samples in client $k$. $N=\sum_{k=1}^{K} N^{(k)}$ is the total number of samples in all clients. The goal of the clients is to jointly learn a global model $f$ with parameters $\theta$ orchestrated by a central server. The global objective function is defined by
\begin{equation}  \label{fl}
    \min\limits_\theta \mathcal{L}(\theta)\coloneqq \sum_{k=1}^{K} \frac{N^{(k)}}{N} \mathcal{L}^{(k)}(\theta),
\end{equation}
where $\mathcal{L}^{(k)}(\theta)=\mathbb{E}_{(x_i^{(k)}, y_i^{(k)}) \sim \mathcal{D}^{(k)}}[\ell(f(x_i^{(k)}; \theta), y_i^{(k)})]$ is the local expected risk function in client $k$. Here, $\ell(\cdot, \cdot)$ denotes the loss function, such as the cross-entropy loss for the classification task and the mean-square error loss for the regression task. In practice, each client usually performs multiple local updates to optimize $\theta$ via Stochastic Gradient Descent (SGD) on the empirical version of $\mathcal{L}^{(k)}(\theta) \coloneqq \frac{1}{N^{(k)}}\sum_{i=1}^{N^{(k)}} \ell(f(x_i^{(k)}; \theta), y_i^{(k)})$ during each round \cite{mcmahan2017fl}.

\section{Method}
In this section, we present our proposed framework FedVN tailored for eliminating the distribution shift issue in FGL. We first formulate the problem setup of FGL with distribution shift. We then provide an overview of FedVN, followed by the details of its three main components: a GNN model, VNs, and an edge generator.

\subsection{Problem Setup}
In this work, we focus on the task of graph-level property prediction in FGL. Given a set of $K$ clients, each client $k$ has its local graph dataset $\mathcal{D}^{(k)}=\{(\mathcal{G}_i^{(k)}, y_i^{(k)})\}_{i=1}^{N^{(k)}}$ drawn from its own graph data distribution $P^{(k)}(\mathcal{G}, y)$.
As the clients may collect their local graph data for different purposes, FGL can suffer from significant distribution shifts, i.e., $P^{(k)}(\mathcal{G}, y) \neq P^{(j)}(\mathcal{G}, y)$ for client $k$ and $j$, while $P^{(k)}(y) = P^{(j)}(y)$. Following many studies in graph out-of-distribution generalization \cite{gui2024joint, sui2024unleashing, zhuang2024imold}, we assume that a given graph $\mathcal{G}$'s label is invariantly determined by its causal substructure $\mathcal{G}_c$. The remaining part of $\mathcal{G}$ is the non-causal substructure denoted by $\mathcal{G}_n$. In our FGL setting, we assume that the non-causal substructure shares a similar pattern across graphs within a client but differs significantly among the clients, i.e., $P^{(k)}(\mathcal{G}_n) \neq P^{(j)}(\mathcal{G}_n)$.
The goal of these clients is to jointly train a GNN model $f$ with parameters $\theta$ for graph property prediction.

\subsection{Proposed Method: FedVN}
To deal with the above problem, we propose a novel FGL framework FedVN in this section. Figure \ref{fig:framework} illustrates an overview of FedVN. The intuition of FedVN is to let the clients manipulate their local graph data through learnable graph augmentation strategies in order that the global GNN model can be trained over identical manipulated graph data without any distribution shift across clients.
To achieve this, the key point is to design a proper scheme for graph augmentation. Inspired by recent studies about VNs \cite{gilmer2017neural, hu2021ogb, hu2020open} in graph learning, we propose to learn graph augmentation with extra VNs to eliminate distribution shifts in FGL. More specifically, the clients in FedVN collaboratively learn a set of shared VNs while training a global GNN model. Considering the cross-client distribution shift, FedVN enables each client to learn a personalized edge predictor that determines how the VNs connect its local graphs. In the following, we will introduce the three components of FedVN in detail.

\subsubsection{GNN model.}
The goal of FedVN is to train a global GNN model over distributed graph data from multiple clients. The GNN model could employ common GNNs like GIN \cite{xu2018gin} as the encoder. Different from existing FGL studies that directly train GNN models over raw graph data, we propose to train a GNN model over the manipulated graph data identical across clients by client-specific graph augmentation strategies. More specifically, given each graph $\mathcal{G}_i^{(k)}$ with its corresponding label $y_i^{(k)}$ in client $k$, the personalized graph augmentation $\phi^{(k)}$ learns to manipulate $\mathcal{G}_i^{(k)}$ into an augmented graph $\tilde{\mathcal{G}}_i^{(k)}$. Therefore, the local empirical risk to optimize $\theta$ in client $k$ can be rewritten as the average supervised loss over augmented graphs by
\begin{equation} \label{loss_s}
\begin{aligned}
    \mathcal{L}_S^{(k)} & = \frac{1}{N^{(k)}}\sum_{i=1}^{N^{(k)}} \ell(f(\tilde{\mathcal{G}}_i^{(k)};\theta), y_i^{(k)}), \\
    & \textrm{where} \; \tilde{\mathcal{G}}_i^{(k)} = \phi^{(k)}(\mathcal{G}_i^{(k)}).\\
\end{aligned}
\end{equation}
In this study, we propose to realize graph augmentation by adding multiple VNs into local graphs to obtain desired graphs for training the global GNN model.

\subsubsection{Virtual nodes.} \label{section: vn}
Adding a VN that uniformly connects to all input graph nodes is an effective graph augmentation technique for graph-level tasks \cite{cai2023connection, gilmer2017neural}. However, the role of VNs introduced in our proposed FedVN is essentially different. VNs in FedVN are designed to eliminate the distribution shift of graph data across clients, while the previous studies use a single VN for modeling long-range dependencies within a graph \cite{cai2023connection} since the VN serves as a shortcut linking every two nodes in the graph. In fact, simply adding a VN may not effectually bridge divergences between graphs due to the following two challenges.
First, a single VN can be inadequate for different graphs. 
Considering the clients in Figure \ref{fig:framework}, we may need at least three types of VNs to obtain the desired identical graphs in the clients. Second, a VN, such as the square VN in Client $k$, is usually not supposed to equally connect all the nodes in an input graph, which is different from what the VN typically does. These two challenges motivate us to conceive a nontrivial mechanism for integrating VNs into our proposed framework.

To deal with the above two challenges, we propose to augment local graphs by learning multiple VNs in FedVN. More concretely, the clients will jointly learn $M$ shared VNs with the feature matrix $\textbf{Q} \in \mathbb{R}^{M\times d_x}$. Each VN $m$ has a unique feature vector $\textbf{q}_m \in \textbf{Q}$ with the same dimension as the original graph nodes. 
Additionally, each client will determine how the $M$ VNs connect their local graphs in a personalized manner. Specifically, for each graph $\mathcal{G}_i^{(k)}$ in client $k$, we inject each VN $m$ into $\mathcal{G}_i^{(k)}$ by assigning a weighted edge with weight $s_{v,m} \in [0, 1]$ between $m$ and each graph node $v$. $s_{v,m}$ measures how significantly $m$ and $v$ impact each other, and they are disconnected when $s_{v,m}=0$. With the VNs, we can rewrite Equation (\ref{gnn}) in a modified version for each node $v \in \mathcal{V}_i^{(k)}$ by
\begin{equation}
\begin{aligned}
    \textbf{h}^{(l)}_v = \textrm{CB}_\text{gn}^{(l)}(\textbf{h}_v^{(l-1)},
    & \textrm{AGG}_{\text{gn}}^{(l)}(\{\textbf{h}_u^{(l-1)}: u \in \mathcal{N}(v)\}))  \\
    & + \sum\limits_{m=1}^M s_{v,m} \cdot \textbf{h}_m^{(l-1)},\\
\end{aligned}
\end{equation}
where
$\textbf{h}_m^{0}$ is the feature vector $\textbf{q}_m$ of VN $m$, and $\mathcal{V}_i^{(k)}$ denotes the node set in $\mathcal{G}_i^{(k)}$. In the meantime, each VN $m$ updates its representation by aggregating information from every graph node $v$ through
\begin{equation}
    \textbf{h}^{(l)}_m = \textrm{CB}_\text{vn}^{(l)}(\textbf{h}_m^{(l-1)}, \textrm{AGG}_{\text{vn}}^{(l)}(\{ s_{v,m} \cdot \textbf{h}_v^{(l-1)}: v \in \mathcal{V}_i^{(k)}\})),
\end{equation}
where $\textrm{AGG}_{\text{vn}}^{(l)}(\cdot)$ and $\textrm{CB}_\text{vn}^{(l)}(\cdot)$ represent the aggregation and combination operations for VNs, respectively.

One potential issue when augmenting local graphs with the VNs is that they are indistinguishable without any prior knowledge. Although the clients are seemingly learning multiple VNs, the VNs are prone to collapse to a single or fewer points (we also observe this collapsing in our empirical experiments). Hence, multiple VNs will be essentially equivalent to fewer VNs and consequently fail to tackle distribution shifts in FGL. To avoid collapsing solutions, we propose to decouple VNs during local training in FedVN. The intuition here is to compel the VNs to reside in the whole feature space as much as possible so that they become dissimilar to each other. We achieve this in FedVN by diminishing the correlation between each pair of VNs.
Specifically, we design a decoupling loss \cite{shi2022feddecorr} during local training with the Frobenius norm of local $\textbf{Q}$'s correlation matrix in each client $k$ by
\begin{equation} \label{loss_v}
    \mathcal{L}_V^{(k)}=\frac{1}{M^2} \Vert \mathbf{\Sigma} \Vert_F^2,
\end{equation}
where $\mathbf{\Sigma}$ denotes the correlation matrix of $\textbf{Q}$, and $\Vert\cdot\Vert_F$ represents the Frobenius norm. 

\subsubsection{Edge generation.}
Given the shared VNs, each client in FedVN is expected to learn a personalized graph augmentation strategy where the shared VNs connect their local graphs in a client-specific fashion so that the distribution shift can be eliminated. 
One straightforward approach is to construct edges based on the feature similarity between graph nodes and VNs. Although this feature similarity-based approach is widely adopted in graph structure learning \cite{chen2020iterative, fatemi2021slaps}, it is unsuitable for this scenario. In fact, edge patterns are significantly impacted by structure information as well. For instance, the house motif \includegraphics[height=0.4cm]{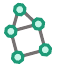} in Figure \ref{fig:framework} causally determines the label of \textit{House}. Therefore, the circle nodes from the causal substructure and the non-causal part should have different patterns to connect VNs even though they have the same node type with identical node features. 

To tackle the above issue, we propose to introduce an extra edge generator $g$ with parameters $\omega$ in FedVN to construct edges between VNs and graph nodes. The edge generator is composed of a GNN encoder and an MLP projector. 
Given a graph $\mathcal{G}_i^{(k)}$ in client $k$, the output of the edge generator is a score matrix $\textbf{S}_i^{(k)} \in \mathbb{R}^{|\mathcal{V}_i^{(k)}|\times M}$. Specifically, the GNN encoder first encodes each node $v$ and its neighboring information to obtain its embedding $\textbf{e}_v$ following Equation (\ref{gnn}). The MLP projector then transforms $\textbf{e}_v$ into an $M$-dimensional score vector $\textbf{s}_v \in \textbf{S}_i^{(k)}$ after a sigmoid function. Each score value $s_{v,m}$ in $\textbf{s}_v$ is adopted as the edge weight between node $v$ and VN $m$.

Despite the prowess of the edge generator in learning edge patterns regarding graph structures, the learned edge patterns cannot be guaranteed to follow a client-specific manner without explicit guidance. Instead, the edge generator may connect VNs and graph nodes within a client very diversely. Note that we assume graphs usually share similar non-causal substructures within a client but have diverse ones across clients (e.g., the function groups - the phenyl group in Institute A and the ester in Institute B - in our previous example). Therefore, the learned edge patterns on the graphs are expected to have similar distributions within a client and differ significantly across clients.
Motivated by this, we propose a novel score-contrastive loss that encourages the total score vector of a graph to get close to the expected one of the graphs within the client and keep away from the expected one of the graphs from other clients. Let $\tilde{\textbf{s}}_i^{(k)}=\sum_{v \in \mathcal{V}_i^{(k)}} \textbf{s}_v$ denote the sum of score vectors in graph $\mathcal{G}_i^{(k)}$. $\textbf{s}_{local}=\frac{1}{N^{(k)}}\sum_{i=1}^{N^{(k)}}\tilde{\textbf{s}}_i^{(k)}$ is the average of $\tilde{\textbf{s}}_i^{(k)}$ in client $k$, and $\textbf{s}_{global}$ denotes the global expected sum of score vectors.
Inspired by NT-Xent loss \cite{chen2020simple, sohn2016improved}, we define $\tilde{\textbf{s}}_i^{(k)}$ and $\textbf{s}_{local}$ as a positive pair while $\tilde{\textbf{s}}_i^{(k)}$ and $\textbf{s}_{global}$ are defined as a negative pair. Therefore, our score-contrastive loss can be formulated as
\begin{equation} \label{loss_e}
    \mathcal{L}_E^{(k)} = -\frac{1}{N^{(k)}}\sum_{i=1}^{N^{(k)}} \log \frac{e^{\text{sim}(\tilde{\textbf{s}}_i^{(k)}, \textbf{s}_{local}) / \tau}}
    {e^{\text{sim}(\tilde{\textbf{s}}_i^{(k)}, \textbf{s}_{local}) / \tau}+e^{\text{sim}(\tilde{\textbf{s}}_i^{(k)}, \textbf{s}_{global}) / \tau}},
\end{equation}
where $\text{sim}(\cdot,\cdot)$ represents the formula of cosine similarity, and $\tau$ is the temperature. Empirically, $\textbf{s}_{global}$ could be simply substituted by a positive constant vector.

Finally, we obtain the local objective function of FedVN in client $k$ formulated by 
\begin{equation}
    \min_{\theta, \textbf{Q}, \omega^{(k)}}  \mathcal{L}_S^{(k)} + \lambda_1 \mathcal{L}_V^{(k)} + \lambda_2 \mathcal{L}_E^{(k)},
\end{equation}
where $\lambda_1$ and $\lambda_2$ are two hyperparameters to balance the impact of loss terms.

\subsection{Algorithmic Design}
Like other FGL frameworks, the proposed FedVN consists of two stages: local training and global update. Each client updates the GNN model, VNs, and the edge generator during local training. Then, the local GNN model and VNs are uploaded to the server for global update. The overall algorithm of FedVN can be found in our technical appendix.

\subsubsection{Local training.}
During local training, we first optimize the local edge generator based on the global GNN model and VNs. We then optimize the GNN model and VNs with the updated edge generator.

\noindent $\blacktriangleright$ \textbf{Step 1: Fix $\theta$ and $\textbf{Q}$, update $\omega^{(k)}$}. Train $\omega^{(k)}$ with the global $\theta$ and $\textbf{Q}$ by
\begin{equation} \label{update_omega}
    \omega^{(k)} \leftarrow \omega^{(k)} - \eta_\omega \nabla_\omega (\mathcal{L}_S^{(k)} + \lambda_2 \mathcal{L}_E^{(k)}).
\end{equation}

\noindent $\blacktriangleright$ \textbf{Step 2: Fix $\omega^{(k)}$, update $\theta^{(k)}$ and $\textbf{Q}^{(k)}$}. After updating $\omega^{(k)}$, train $\theta^{(k)}$ and $\textbf{Q}^{(k)}$ by
\begin{equation} \label{update_theta}
    \theta^{(k)} \leftarrow \theta^{(k)} - \eta_\theta \nabla_\theta \mathcal{L}_S^{(k)},
\end{equation}
\begin{equation} \label{update_q}
    \textbf{Q}^{(k)} \leftarrow \textbf{Q}^{(k)} - \eta_Q \nabla_\textbf{Q} (\mathcal{L}_S^{(k)} + \lambda_1 \mathcal{L}_V^{(k)}).
\end{equation}
$\eta_\omega$, $\eta_\theta$, and $\eta_Q$ are their corresponding learning rates.

\subsubsection{Global update.}
The server updates the GNN model and VNs following FedAvg \cite{mcmahan2017fl}. Specifically, the global GNN model and VNs are updated as the weighted average of their local version by
\begin{equation} \label{global_update}
    \theta = \sum_{k=1}^{K} \frac{N^{(k)}}{N} \theta^{(k)},\; \; \textbf{Q} = \sum_{k=1}^{K} \frac{N^{(k)}}{N} \textbf{Q}^{(k)}
\end{equation}
The complexity analysis of FedVN is provided in our technical appendix due to the page limit.

\begin{table*}[]
\centering
\label{table:main}
\setlength\tabcolsep{15pt}
\begin{tabular}{cccccccc}
\toprule
Dataset        & \multicolumn{2}{c}{Motif}  & CMNIST     & ZINC  & SST2    \\
\cmidrule(lr){1-1}\cmidrule(lr){2-3}\cmidrule(lr){4-4}\cmidrule(lr){5-5}\cmidrule(lr){6-6}
Metric  & \multicolumn{2}{c}{Accuracy $\uparrow$}  & Accuracy $\uparrow$     & MAE $\downarrow$   & Accuracy $\uparrow$    \\
\cmidrule(lr){1-1}\cmidrule(lr){2-3}\cmidrule(lr){4-4}\cmidrule(lr){5-5}\cmidrule(lr){6-6}
Partition setting  & Basis             & Size              & Color             & Scaffold              & Length  \\ 
\cmidrule(lr){1-1}\cmidrule(lr){2-3}\cmidrule(lr){4-4}\cmidrule(lr){5-5}\cmidrule(lr){6-6}
Self-training   &  67.12$\pm$0.89   &  47.60$\pm$2.32   &  39.38$\pm$0.90   &  0.5442$\pm$0.0146    &  80.54$\pm$0.67 \\ 
FedAvg          &  58.70$\pm$2.39   &  47.82$\pm$3.16   &  39.18$\pm$0.92   &  0.6235$\pm$0.0158    &  81.79$\pm$0.27 \\ 
FedProx         &  57.90$\pm$1.36   &  47.88$\pm$4.08   &  39.78$\pm$0.68   &  0.6235$\pm$0.0165    &  81.74$\pm$0.33 \\  
\cmidrule(lr){1-1}\cmidrule(lr){2-3}\cmidrule(lr){4-4}\cmidrule(lr){5-5}\cmidrule(lr){6-6}
FedBN           &  58.44$\pm$1.33   &  47.54$\pm$2.66   &  39.26$\pm$0.76   &  0.5129$\pm$0.0119    &  81.73$\pm$0.35\\ 
Ditto           &  63.38$\pm$0.89   &  47.48$\pm$3.20   &  39.00$\pm$0.94   &  0.5471$\pm$0.0146    &  81.69$\pm$0.67 \\
FedRep          &  59.20$\pm$2.83   &  45.48$\pm$0.86   &  36.78$\pm$0.67   &  0.5220$\pm$0.0110    &  74.77$\pm$2.84 \\
FedALA          &  59.92$\pm$1.14   &  48.52$\pm$3.34   &  39.22$\pm$1.12   &  0.5837$\pm$0.0159    &  81.77$\pm$0.61 \\
GCFL+           &  57.36$\pm$2.00   &  49.34$\pm$2.70   &  38.82$\pm$1.11   &  0.6224$\pm$0.0147   &  81.39$\pm$0.45 \\
FedStar         &  63.62$\pm$4.85   &  45.68$\pm$2.11   &  28.10$\pm$1.17   &  0.5963$\pm$0.0163    &  58.57$\pm$1.25 \\ 
\cmidrule(lr){1-1}\cmidrule(lr){2-3}\cmidrule(lr){4-4}\cmidrule(lr){5-5}\cmidrule(lr){6-6}
\textbf{FedVN (Ours)}       &  \textbf{75.72$\pm$1.85}   &  \textbf{50.41$\pm$1.17}   &  \textbf{43.67$\pm$1.25}  &  \textbf{0.4947$\pm$0.0174}    &  \textbf{83.13$\pm$0.79} \\ \bottomrule
\end{tabular}
\caption{Performance of FedVN and other baselines over four datasets under five settings. }

\end{table*}

\section{Theoretical Analysis}
In this section, we provide theoretical analyses for our proposed FedVN. As discussed above, FedVN enables each client to eliminate the distribution shift through a personalized edge generator determining how the VNs connect local graphs in a client-specific manner. It means that we can obtain identical representations of augmented graphs simply by desired score matrices. To simplify expression, we consider a part $\theta_e$ in $f$ to generate graph embeddings (i.e., a GNN model except for its prediction head).

\noindent \textbf{Theorem 1.} \textit{Consider a pair of graphs $\mathcal{G}$ and $\mathcal{G}'$. For an arbitrary feature matrix} $\textbf{Q}\in \mathbb{R}^{M\times d_x}$ \textit{with} $M = d_x$ \textit{and any given GNN model $f$'s component $\theta_e$, when} $\textbf{Q}$ \textit{has full rank, there exists a pair of score matrices} $\textbf{S}$ \textit{and} $\textbf{S}'$ \textit{that satisfies: }
\begin{equation}
\begin{aligned}
    f(\tilde{\mathcal{G}};\theta_e) & = f(\tilde{\mathcal{G}'};\theta_e), \\ 
    \text{where} \;  \tilde{\mathcal{G}} = \phi(\mathcal{G}, \textbf{S}, \textbf{Q}) \; & \text{and} \; \tilde{\mathcal{G}'} = \phi(\mathcal{G}', \textbf{S}', \textbf{Q}). \\ 
\end{aligned}
\end{equation}

The complete proof of Theorem 1 is provided in our technical appendix. According to Theorem 1, we can conclude that FedVN can obtain identical graph embedding of $\mathcal{G}$ and $\mathcal{G}'$ given any $\textbf{Q}$ and $\theta_e$. Therefore, FedVN has the capability of eliminating the distribution shift issue in FGL.

Note that we assume $\textbf{Q}$ has full rank in Theorem 1. This requires that multiple VNs cannot collapse to fewer nodes. To avoid collapsing VNs, we design a decoupling loss in FedVN. Formally, we can conclude the following theorem.

\noindent \textbf{Theorem 2.} \textit{Minimizing $\mathcal{L}_V^{(k)}$ drives} $\textbf{Q}$ \textit{to have full rank.}

The complete proof of Theorem 2 is provided in our technical appendix. As a result, the learned VNs are distinct from each other, avoiding VN collapse.

\section{Experiments}
\subsection{Experimental Setup}
\subsubsection{Datasets.} 
We adopt graph datasets in \cite{gui2022good} to simulate distributed graph data in multiple clients. Specifically, we use four datasets, including Motif, CMNIST, ZINC, and SST2. We split each graph dataset into multiple clients according to its environment settings so that every client has local graphs from one environment. More details about these datasets can be found in our technical appendix.

\subsubsection{GNN backbones.}
We follow previous studies \cite{xie2021gcfl, tan2023fedstar} to choose GNN backbones. Specifically, we adopt a three-layer GIN as the encoder and a two-layer MLP as the prediction head in the GNN model, with \textit{mean} pooling as the READOUT operation. The edge predictor similarly uses a three-layer GIN as the encoder, followed by a two-layer MLP as the projector.

\subsubsection{Baselines.}
We adopt nine baselines in our experiments. Among them, FedAvg \cite{mcmahan2017fl} and FedProx \cite{li2020fedprox} are two typical FL frameworks for learning a global model. As for personalized FL frameworks, we use
FedBN \cite{li2020fedbn}, Ditto \cite{li2021ditto}, FedRep \cite{collins2021fedrep}, FedALA \cite{zhang2023fedala}. 
Furthermore, we also include two recent approaches for graph property prediction in FGL, i.e., GCFL+ \cite{xie2021gcfl} and FedStar \cite{tan2023fedstar}. More details of the baselines are provided in our technical appendix.

\subsubsection{Implementation details.}
We run our experiment for five random repetitions and report the average results in the experiments.  Each repetition runs 100 epochs. The local epoch is set to 1, and the batch size is 32. The hidden size of the GNN model and the edge predictor is set to 100. We use SGD as the optimizer for local updates with a learning rate set to 0.01 for CMNIST and 0.001 for others. The temperature $\tau$ is set to 0.1.
More information about hyperparameter settings can be found in our technical appendix.

\subsection{Comparative Results}

\subsubsection{Effectiveness of FedVN.}
We first evaluate the overall performance of FedVN and other baselines. Table 1 shows the results of FedVN and baselines over the datasets. From Table 1, we can observe that our proposed FedVN can consistently outshine the baselines on the five tasks. Specifically, FedVN achieves about accuracy improvement of 8\% on Motif/basis. Since Motif/basis is a synthetic dataset where graphs in a client all have the same label-irrelevant base subgraph, FedVN can eliminate such distribution shifts by adding VNs to local graphs. As for another semi-synthetic dataset, FedVN can perform well on CMNIST/Color. Considering the distribution shift in CMNIST/Color mainly on node features, the results indicate that FedVN can still overcome distribution shifts in terms of node features. Furthermore, FedVN can also outperform other baselines on the two real-world datasets ZINC/Scaffold and SST2/Length.

\subsubsection{Convergence speed.}
\begin{figure}[!t]
\setlength {\belowcaptionskip} {-0.3cm}
\centering
\includegraphics[width=\linewidth]{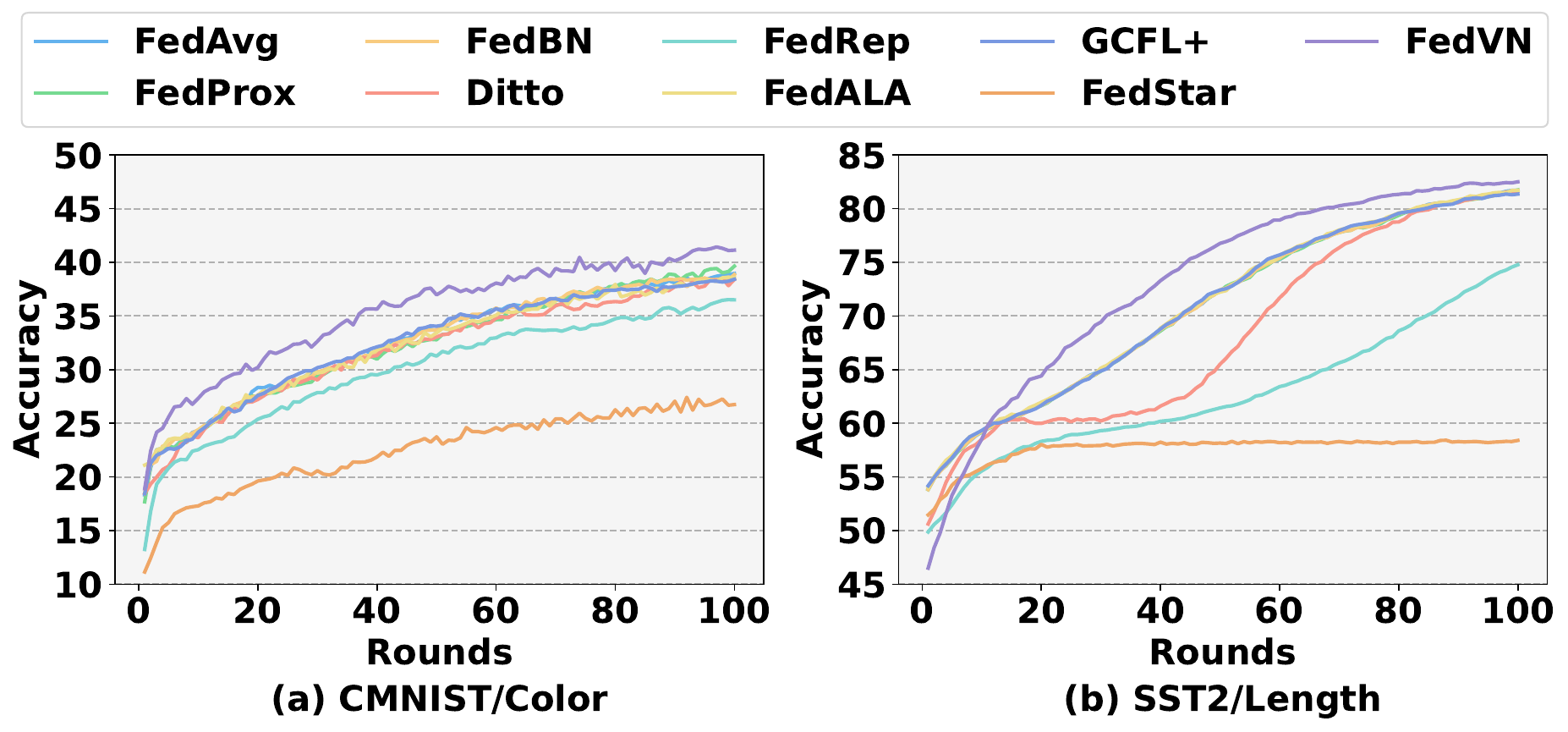}
\caption{Convergence curves of FedVN and other baselines on CMNIST/Color and SST2/Length.}
\label{fig:convergence}
\end{figure}

Figure \ref{fig:convergence} illustrates the accuracy curves of FedVN and other baselines on CMNIST/Color and SST2/Length. We can observe that our method can outperform other baselines during the training process. Besides, we notice that FedStar suffers from significant performance degradation on CMNIST/Color and SST/Length. We argue that graphs in the two datasets are mainly labeled according to their node features. In this case, sharing structure encoders may not help enhance model utility.

\subsection{Analysis of FedVN}
\subsubsection{Influence of $\lambda_1$ and $\lambda_2$.}
We conduct experiments to explore the sensitivities of two hyperparameters $\lambda_1$ and $\lambda_2$. Figure \ref{fig:heatmap_bar}(a) presents the accuracy heatmap of FedVN on SST2/Length with different values of $\lambda_1$ and $\lambda_2$. We observe that FedVN achieves good performance when $\lambda_1$ is 0.1$\sim$1 and $\lambda_2$ is 1$\sim$10. The performance of FedVN will degrade apparently if either $\lambda_1$ or $\lambda_2$ is 0. The above observation demonstrates that the two loss terms $\mathcal{L}_V^{(k)}$ and $\mathcal{L}_E^{(k)}$ are necessary in FedVN.

\subsubsection{Influence of VN numbers.}
\begin{figure}[!t]
\centering
\includegraphics[width=\linewidth]{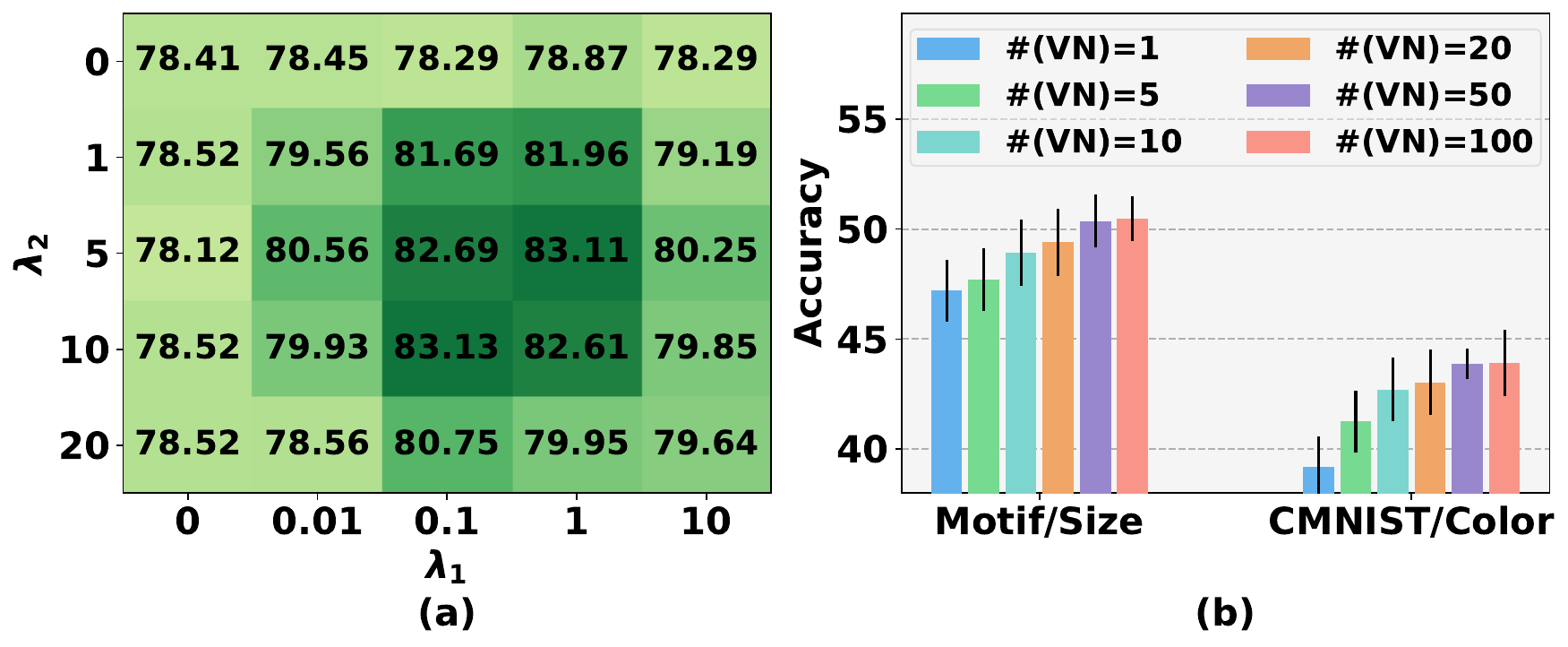}
\caption{(a) Performance of FedVN on SST2/Length with different values of $\lambda_1$ and $\lambda_2$. (b) Performance of FedVN with different numbers of VNs.}
\label{fig:heatmap_bar}
\end{figure}

As discussed in Section \ref{section: vn}, too few VNs are usually insufficient to eliminate cross-client distribution shifts. In contrast, a large number of VNs can bring extra communication and computational costs. We investigate the performance of fedVN with different numbers of VNs and report the results on Motif/Size and CMNIST/Color in Figure \ref{fig:heatmap_bar}(b). From the figure, we can observe that FedVN can achieve good performance consistently when equipped with enough VNs. When there are only a few VNs (e.g., 1 or 5), the performance of FedVN is limited. This observation validates our motivation for involving multiple VNs in FedVN. In addition, since too many VNs (e.g., when $M=d_x=100$ in our experiments) will be very hard to train, we may consider using fewer VNs in practice.

\subsubsection{Effectiveness of edge generators.}
One insight in our study is personalized edge generators to learn client-specific graph augmentation strategies for eliminating distribution shifts of graph data. Without edge generators, multiple VNs will uniformly connect graph nodes in local graphs. In this case, these VNs can be merged into a single VN, which can be regarded as an equivalent scheme in previous studies \cite{hu2020open, hu2021ogb}.
To validate the effectiveness of edge generators in FedVN, we compare the performance of FedVN and its degraded version without edge generators. We report the results on Motif/Basis and CMNIST/Color in Table 2. We observe that FedVN without edge generators encounters apparent performance degradation, which validates the necessity of edge generators in FedVN.

\subsubsection{More experimental results.} Due to the page limit, more experimental results are provided in our technical appendix, including visualizations of VN collapse, visualizations of distribution shifts in FedAvg and FedVN.

\begin{table}[]
\centering
\label{table:without_g1}
\setlength\tabcolsep{10pt}
\begin{tabular}{lccccccc}
\toprule
Dataset         & Motif/Basis $\uparrow$   & CMNIST/Color $\uparrow$  \\  \midrule 
FedVN           &  75.72$\pm$1.85          & 43.67$\pm$1.25             \\ 
FedVN w/o $g$   &  60.20$\pm$1.75          & 40.15$\pm$1.27        \\ \bottomrule
\end{tabular}
\caption{Performance of FedVN and its variant without edge generators on Motif/Basis and CMNIST/Color.}

\end{table}

\section{Conclusion}
In this study, we take the first step towards investigating the distribution shift issue of graph data across clients for graph property prediction in FGL. We grapple with this issue by designing personalized graph augmentation to eliminate cross-client distribution shifts in FGL. We propose FedVN, a novel FGL framework that integrates multiple learnable VNs into local graphs in a client-specific fashion. To achieve this, each client is equipped with a personalized edge generator which determines how the VNs connect its local graphs. In this way, the clients can jointly train a powerful global GNN model over augmented graph data identical across clients. Furthermore, we provide theoretical analyses to validate that our design in FedVN has the capability of eliminating distribution shifts in FGL. Our extensive experiments show that FedVN outshines nine baselines on five tasks.

\section*{Acknowledgments}
This work is supported in part by the National Science Foundation under grants IIS-2006844, IIS-2144209, IIS-2223769, IIS-2331315, CNS-2154962, BCS-2228534, and CMMI-2411248 and the Commonwealth Cyber Initiative Awards under grants VV-1Q24-011, VV-1Q25-004.

\bibliography{aaai25}

\appendix
\section{Proof of Theorem 1} \label{appendix:theorem 1}
\noindent \textbf{Theorem 1.} \textit{Consider a pair of graphs $\mathcal{G}$ and $\mathcal{G}'$. For an arbitrary feature matrix} $\textbf{Q}\in \mathbb{R}^{M\times d_x}$ \textit{with} $M = d_x$ \textit{and any given GNN model $f$'s component $\theta_e$, when} $\textbf{Q}$ \textit{has full rank, there exists a pair of score matrices} $\textbf{S}$ \textit{and} $\textbf{S}'$ \textit{that satisfies: }
\begin{equation}
\begin{aligned}
    f(\tilde{\mathcal{G}};\theta_e) & = f(\tilde{\mathcal{G}'};\theta_e), \\ 
    \text{where} \;  \tilde{\mathcal{G}} = \phi(\mathcal{G}, \textbf{S}, \textbf{Q}) \; & \text{and} \; \tilde{\mathcal{G}'} = \phi(\mathcal{G}', \textbf{S}', \textbf{Q}). \\ 
\end{aligned}
\end{equation}

\begin{proof}
To prove Theorem 1, we first consider a specific architecture of the GNN encoder in $f$. Specifically, we instantiate the GNN encoder as a single-layer GIN \cite{xu2018gin} with a linear transformation. We then extend our proof to other GNN layers and multi-layer GNN encoders.

\subsubsection{A single-layer GIN as the GNN encoder.} Given a graph $\mathcal{G}$ with its adjacency matrix $\textbf{A}$, a single-layer GIN with $sum$ pooling computes its embedding with the added VNs by
\begin{equation}
    \tilde{\textbf{h}}= f(\tilde{\mathcal{G}};\theta_e) = \textbf{1}^\top \cdot[(\textbf{A} + (1+\epsilon) \cdot \textbf{I})\cdot \textbf{X} \cdot \textbf{W} + \textbf{S} \cdot \textbf{Q}].
\end{equation}
where $\textbf{1}$ is a constant vector with each entry equal to 1, and $\textbf{I}$ denotes an identity matrix. $\epsilon$ and $\textbf{W}$ are learnable parameters in $\theta_e$. Let $\textbf{H} = (\textbf{A} + (1+\epsilon) \cdot \textbf{I})\cdot \textbf{X} \cdot \textbf{W}$. We have 
\begin{equation}
    \tilde{\textbf{h}}= f(\tilde{\mathcal{G}};\theta_e) = \textbf{1}^\top \cdot[\textbf{H} + \textbf{S} \cdot \textbf{Q}],
\end{equation}
To obtain an identical embedding of the two graphs with VNs, we need
\begin{equation} \label{equation:18}
    \textbf{1}^\top_{|\mathcal{V}|} \cdot[\textbf{H} + \textbf{S} \cdot \textbf{Q}] = \textbf{1}^\top_{|\mathcal{V}'|} \cdot[\textbf{H}' + \textbf{S}' \cdot \textbf{Q}].
\end{equation}
Given that $\textbf{Q}$ has full rank, there exists an inverse matrix of $\textbf{Q}$, denoted as $\textbf{Q}^r$. We can rewrite Equation (\ref{equation:18}) by 
\begin{equation}
    \textbf{1}^\top_{|\mathcal{V}|}  \cdot[\textbf{H} \cdot \textbf{Q}^r + \textbf{S}] = \textbf{1}^\top_{|\mathcal{V}'|}  \cdot[\textbf{H}' \cdot \textbf{Q}^r + \textbf{S}'].
\end{equation}
Then, we have
\begin{equation} \label{equation:20}
    \textbf{1}^\top_{|\mathcal{V}'|} \cdot \textbf{S}' = \textbf{1}^\top_{|\mathcal{V}|} \cdot \textbf{S} +(\textbf{1}^\top_{|\mathcal{V}|}  \cdot \textbf{H} - \textbf{1}^\top_{|\mathcal{V}'|}  \cdot \textbf{H}') \cdot \textbf{Q}^r.
\end{equation}
Therefore, the two graphs will obtain identical graph embedding when $\textbf{S}$ and $\textbf{S}'$ satisfy Equation (\ref{equation:20}).

\subsubsection{Extension to other GNN backbones.} 
Compared with a GIN layer, other GNN backbones (e.g., GCN \cite{kipf2016gcn}) may use different mechanisms to compute $H$. Following \cite{fang2024universal}, we abstract various GNN backbones by 
\begin{equation}
    \textbf{H} = \textbf{D} \cdot \textbf{X} \cdot \textbf{W},
\end{equation}
where $\textbf{D} \in \mathbb{R}^{|\mathcal{V}|\times |\mathcal{V}|} $ is the diffusion matrix. Various GNN backbones adopt different $\textbf{D}$. For instance, $(\textbf{A} + (1+\epsilon) \cdot \textbf{I})$ is the diffusion matrix in GIN. We can recompute $\textbf{H}$ and $\textbf{H}'$ in Equation (\ref{equation:20}) with specific $\textbf{D}$ and $\textbf{D}'$ for any GNN backbone.

\subsubsection{Extension to multi-layer GNNs.} 
After the first GNN layer, we compute the embedding of VNs by 
\begin{equation}
    \textbf{H}_M = \textbf{S}^\top \cdot \textbf{X} \cdot \textbf{W}_M,
\end{equation}
where $\textbf{W}_M$ are learnable parameters in $\theta_e$. As discussed above, $\textbf{S}$ and $\textbf{S}'$ can lead to consistent graph embedding of $\mathcal{G}$ and $\mathcal{G}'$ after a single-layer GNN with shared $\textbf{Q}$. Now we consider a special case where $\textbf{H}_M = \textbf{H}'_M$. In this case, any VN added to $\mathcal{G}$ has the same embedding as that to $\mathcal{G}'$. If we rewrite $\textbf{H}_M$ and $\textbf{H}'_M$ as a single matrix $\textbf{Q}'$, we can regard the second GNN layer as a repetition of the first layer using a new $\textbf{Q}'$. This process can iterate along with the forward pass in multi-layer GNNs and obtain identical graph embedding for the two graphs $\mathcal{G}$ and $\mathcal{G}'$.
\end{proof}

\section{Proof of Theorem 2} \label{appendix:theorem 2}
Before we prove Theorem 2, we first prove the following lemma.

\noindent \textbf{Lemma 1.} \textit{ Given a correlation matrix $\mathbf{\Sigma} \in \mathbb{R}^{M\times M}$ with its singular values $\mathcal{P}=\{p_1, p_2, \cdots, p_M\}$, we have:}
\begin{equation}
    \Vert \mathbf{\Sigma} \Vert_F^2 = M \cdot \text{Var}(\mathcal{P}) + M
\end{equation}
\textit{where} Var\textit{$(\cdot)$ computes the variance of the entries in a set.}


\begin{proof}
    Since $\mathbf{\Sigma}$ is a correlation matrix, it is also a symmetric positive definite matrix. Therefore, the sum of all singular values is equal to its trace. In addition, we know that the diagonal entries in $\mathbf{\Sigma}$ are equal to 1. Then we have
\begin{equation} \label{trace}
    \sum_{m=1}^M p_m = \text{trace}(\mathbf{\Sigma}) = M.
\end{equation}
With Equation (\ref{trace}), we have 
\begin{equation}
\begin{aligned}
    M \cdot \text{Var}(\mathcal{P}) + M
=   & \sum_{m=1}^M(p_m - \frac{1}{M} \sum_{n=1}^N p_n)^2 + M \\
=   & \sum_{m=1}^M(p_m - 1)^2 + M \\
=   & \sum_{m=1}^M p_m^2 - 2\sum_{m=1}^M p_m + 2M \\
=   & \sum_{m=1}^M p_m^2 - 2 M+ 2M \\
=   & \sum_{m=1}^M p_m^2.
\end{aligned}
\end{equation}
Let $\mathbf{\Sigma}=\textbf{U} \mathbf{\Omega}\textbf{V}^\top$ denote the singular value decomposition on $\mathbf{\Sigma}$. We have
\begin{equation}
\begin{aligned}
    \Vert \mathbf{\Sigma} \Vert_F^2
=   & \text{ trace}(\mathbf{\Sigma}^\top \mathbf{\Sigma}) \\
=   & \text{ trace}((\textbf{U} \mathbf{\Omega}\textbf{V}^\top)^\top \textbf{U} \mathbf{\Omega}\textbf{V}^\top) \\
=   & \text{ trace}(\textbf{V}\mathbf{\Omega}^\top \textbf{U}^\top \textbf{U} \mathbf{\Omega}\textbf{V}^\top) \\
=   & \text{ trace}(\textbf{V}\mathbf{\Omega}^\top \mathbf{\Omega}\textbf{V}^\top) \\
=   & \sum_{m=1}^M p_m^2.
\end{aligned}
\end{equation}
Therefore, we have $\Vert \mathbf{\Sigma} \Vert_F^2 = M \cdot \text{Var}(\mathcal{P}) + M$.
\end{proof}

Now, we move to the proof of Theorem 2.

\noindent \textbf{Theorem 2.} \textit{Minimizing $\mathcal{L}_V^{(k)}$ drives} $\textbf{Q}$ \textit{to have full rank.}
\begin{proof}
According to Lemma 1, we can conclude that minimizing the Frobenius norm of $\mathbf{\Sigma}$ is equivalent to minimizing the variance of $\mathbf{\Sigma}$'s singular values. By minimizing the variance of $\mathbf{\Sigma}$'s singular values, we can avoid $p_m = 0$ for any $p_m \in \mathcal{P}$. With all $p_m > 0$, we can get $rank(\textbf{Q})=M$. Therefore, we can conclude that minimizing $\mathcal{L}_V^{(k)}$ drives $\textbf{Q}$ to have full rank.
\end{proof}

\section{Overall Algorithm of FedVN} \label{appendix:algorithm}
The overall algorithm of FedVN is provided in Algorithm \ref{alg:algorithm}.

\begin{algorithm}[!t]
    \caption{Overall algorithm of FedVN}
    \label{alg:algorithm}
    \raggedright
    
    \textbf{Input}: initial global $\theta$, initial global $\textbf{Q} = \textbf{0}$, personalized $\omega^{(k)}$ for each client $k$, \\
    \begin{algorithmic} [1] 
    \FOR{each round $r=1, \cdots, R$}
    \FOR{each client $k$ \textbf{in parallel}}
    \STATE $\theta^{(k)}, \textbf{Q}^{(k)}\leftarrow$ LocalUpdate $(\theta, \textbf{Q})$
    \ENDFOR
    \STATE Update $\theta$ and $\textbf{Q}$ using Equation (13)
    \ENDFOR
    \end{algorithmic}
    \vspace{0.5em}
    
    \textbf{LocalUpdate}$(\theta, \textbf{Q})$: \\
    \begin{algorithmic}[1]
    \STATE \textcolor{gray}{================ Phase 1 =================}
    \FOR{$t=1, \cdots, E$}
    \FOR{each batch of graphs $\mathcal{B} \in \mathcal{D}^{(k)}$}
    \STATE Compute $\mathcal{L}_S^{(k)}$ by Equation (4) and $\mathcal{L}_E^{(k)}$ by Equation (8) with $\theta, \textbf{Q}$
    \STATE Update local edge generator by $\omega^{(k)} \leftarrow \omega^{(k)} - \eta_\omega \nabla_\omega (\mathcal{L}_S^{(k)} + \lambda_2 \mathcal{L}_E^{(k)})$
    \ENDFOR 
    \ENDFOR 
    \STATE \textcolor{gray}{================= Phase 2 ================}
    \STATE $\theta^{(k)}, \textbf{Q}^{(k)} = \theta, \textbf{Q}$
    \FOR{$t=1, \cdots, E$}
    \FOR{each batch of graphs $\mathcal{B} \in \mathcal{D}^{(k)}$}
    \STATE Compute $\mathcal{L}_S^{(k)}$ by Equation (4) and $\mathcal{L}_V^{(k)}$ by Equation (7)
    
    \STATE Update local GNN model by $\theta^{(k)} \leftarrow \theta^{(k)} - \eta_\theta \nabla_\theta \mathcal{L}_S^{(k)}$
    \STATE Update local VNs by $\textbf{Q}^{(k)} \leftarrow \textbf{Q}^{(k)} - \eta_Q \nabla_\textbf{Q} (\mathcal{L}_S^{(k)} + \lambda_1 \mathcal{L}_V^{(k)})$
    \ENDFOR
    \ENDFOR
    \RETURN{$\theta^{(k)}, \textbf{Q}^{(k)}$}
    \end{algorithmic}
\end{algorithm}

\begin{table*}[]
\small
\centering
\caption{Basic information and statistics of graph datasets adopted in our experiments. }
\label{table:dataset}
\setlength\tabcolsep{10.5pt}
\begin{tabular}{lccccccc}
\toprule
\multirow{2}{*}{Dataset}    & \multicolumn{2}{c}{Motif}     & CMNIST        & ZINC          & SST2   \\
\cmidrule(lr){2-3}\cmidrule(lr){4-4}\cmidrule(lr){5-5}\cmidrule(lr){6-6}
                            & Basis         & Size          & Color         & Scaffold      & Length  \\ \midrule
Data type                   & \multicolumn{2}{c}{Synthetic} &  Synthetic    & Molecule      &  Sentence \\
\#(Clients)                 &  5            &  5            &  5            &  10           &  7 \\ 
\#(Graphs)/client           & 1,000         & 1,000         & 1,000         & 1,000         & 1,000\\ 

Task                        & \multicolumn{2}{c}{Classification}   &  Classification   &  Regression &  Classification  \\ 
Metric                      & \multicolumn{2}{c}{Accuracy}   &  Accuracy   &  MAE           &  Accuracy  \\ \bottomrule
\end{tabular}
\end{table*}

\section{Complexity Analysis of FedVN} 
The extra computational cost in FedVN is mainly caused by computing score vectors and VN aggregation. If we consider a 2-layer GCN model with hidden size $d_h$ as the backbone, its computational complexity is approximate $O(d_x d_h |\mathcal{V}|+d_x|\mathcal{E}|)$ for graph $\mathcal{G}=(\mathcal{V}, \mathcal{E}, \mathbf{X})$. Since the edge generator adopts a similar GNN architecture with hidden size $d_e$ to obtain score vectors, the extra computational complexity is also around $O(d_x d_e |\mathcal{V}|+d_x|\mathcal{E}|)$. As for VN aggregation, extra computational complexity $O(M d_h |\mathcal{V}|)$ is required for both Equation (5) and Equation (6). Therefore, the overall extra computational complexity in FedVN is $O(d_x d_e |\mathcal{V}|+d_x|\mathcal{E}|+2M d_h |\mathcal{V}|)$. In practice, we may reduce the computational complexity by setting a smaller $d_e \leq d_h$ or using fewer VNs.

\section{More Details about Experimental Setup}

\subsection{Datasets} \label{appendix:datasets}
We use four graph datasets adapted from \cite{gui2022good}, including Motif, CMNIST, ZINC, and SST2. We present detailed information on these datasets as follows.

\begin{itemize}
    \item \textbf{Motif} is a synthetic dataset designed for structure shifts \cite{wu2021discovering}. Each graph in the dataset is generated by connecting a base graph and a motif, and its label is solely determined by the motif. Graphs are generated using five label-irrelevant base graphs (wheel, tree, ladder, star, and path) and three label-related motifs (house, cycle, and crane). In our experiments, we consider two settings, i.e., partitioning graphs by 1) the base type and 2) graph size.

    \item \textbf{CMNIST} is a semi-artificial dataset designed for node feature shifts. It contains Graphs that are hand-written digits transformed from MNIST using superpixel techniques. Graphs in each client represent digits associated with a client-specific color. We use color as the partition setting for CMNIST.

    \item \textbf{ZINC} is a real-world molecule graph dataset for molecular property regression from ZINC database \cite{gomez2018zinc}. The inputs are molecular graphs in which nodes are atoms, and edges are chemical bonds. The task is to predict the constrained solubility of molecules. In our experiments, we consider partitioning graphs by scaffold, which is the two-dimensional structural base of a molecule.
    
    \item \textbf{SST2} is a real-world natural language sentimental analysis dataset \cite{yuan2022sst}. Each sentence is transformed into a grammar tree graph, where each node represents a word with corresponding word embeddings as node features. The task is to predict the sentiment polarity of a sentence. We select sentence lengths as the partition setting to split graphs.
\end{itemize}
For each setting, we split graphs into multiple clients. Each client retains 1,000 graphs. We summarize basic information and statistics of the datasets in Table \ref{table:dataset}. Among the graphs, 80\% are used for training and the rest for testing. We record the best results of the test set for reporting in the experiment. Therefore, our curve figures showing average values during each round might be different from the best results in the tables.

\subsection{Baselines} \label{appendix:baselines}
We use nine SOTA baselines in our experiments. We provide the details of these baselines as follows.

\begin{itemize}
    \item Self-training: each client trains its own GNN model individually on its local graph data without any communication with each other.

    \item FedAvg \cite{mcmahan2017fl}: the most common FL framework that simply aggregates local models to update the global model.

    \item FexProx \cite{li2020fedprox}: an additional loss term is added to restrict the distance between local models and the global model.

    \item FedBN \cite{li2020fedbn}: batch normalization layers are kept locally while the remaining part is shared across clients.

    \item Ditto \cite{li2021ditto}: an extra personalized model is equipped for each client. On the top of FedAvg, the personalized model is trained to get close to the global model during local training.

    \item FedRep \cite{collins2021fedrep}: only the GNN encoder is shared across clients. Each client learns their own projection head.

    \item FedALA \cite{zhang2023fedala}: each client adaptively aggregates the global model and its local model towards its local objective to initialize the local model before local training.

    \item GCFL+ \cite{xie2021gcfl}: the clients are grouped based on their local GNN model parameters. The clients within a group share the same model parameters.

    \item FedStar \cite{tan2023fedstar}: the feature-structure decoupled GNN model encodes feature information and structure information in a separate manner. Each client learns a personalized feature encoder and jointly learns a global structure encoder with other clients.
\end{itemize}

\subsection{Hyperparameter Settings} \label{appendix:hyperparameter}
$\lambda_1$, $\lambda_2$ and $M$ are three important hyperparameters in our proposed FN. Therefore, we conduct experiments for grid search of the three hyperparameters. Specifically, the search space of each hyperparameter is shown as follows.
\begin{itemize}
    \item $\lambda_1$: 0, 0.01, 0.1, 1, 10
    \item $\lambda_2$: 0, 1, 5, 10, 20
    \item $M$: 1, 5, 10, 20, 50, 100
\end{itemize}

It is worthwhile to note that when we set $M=1$, $\mathcal{L}_V^{(k)}$ will not take effect since it is designed to avoid collapsing of multiple VNs.

\begin{figure}[!t]
\centering
\includegraphics[width=\linewidth]{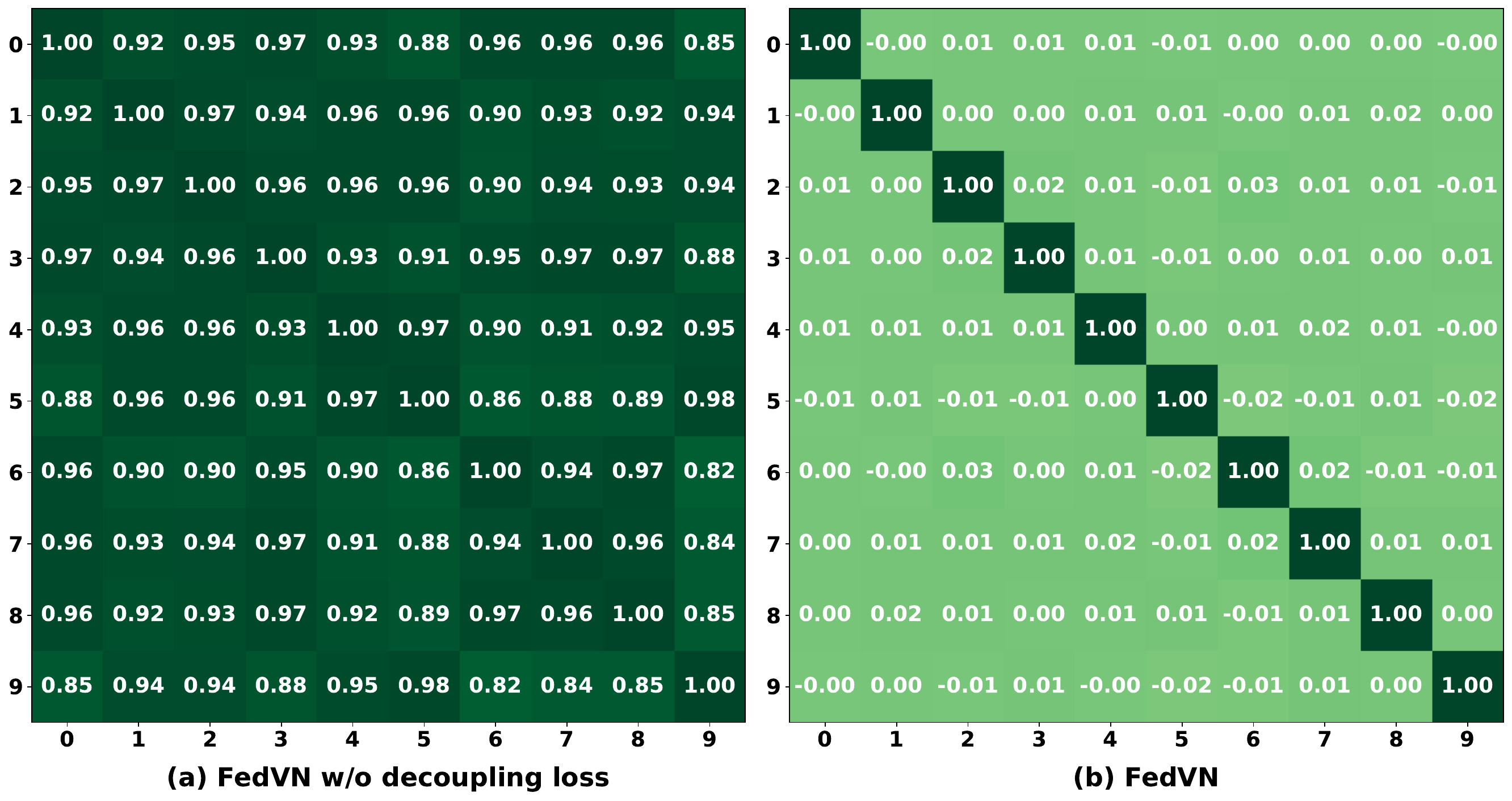}
\caption{Pairwise cosine similarities between 10 VNs on Zinc/Scaffold.}
\label{fig:similarity_zinc}
\end{figure}

\begin{figure}[!t]
\centering
\includegraphics[width=\linewidth]{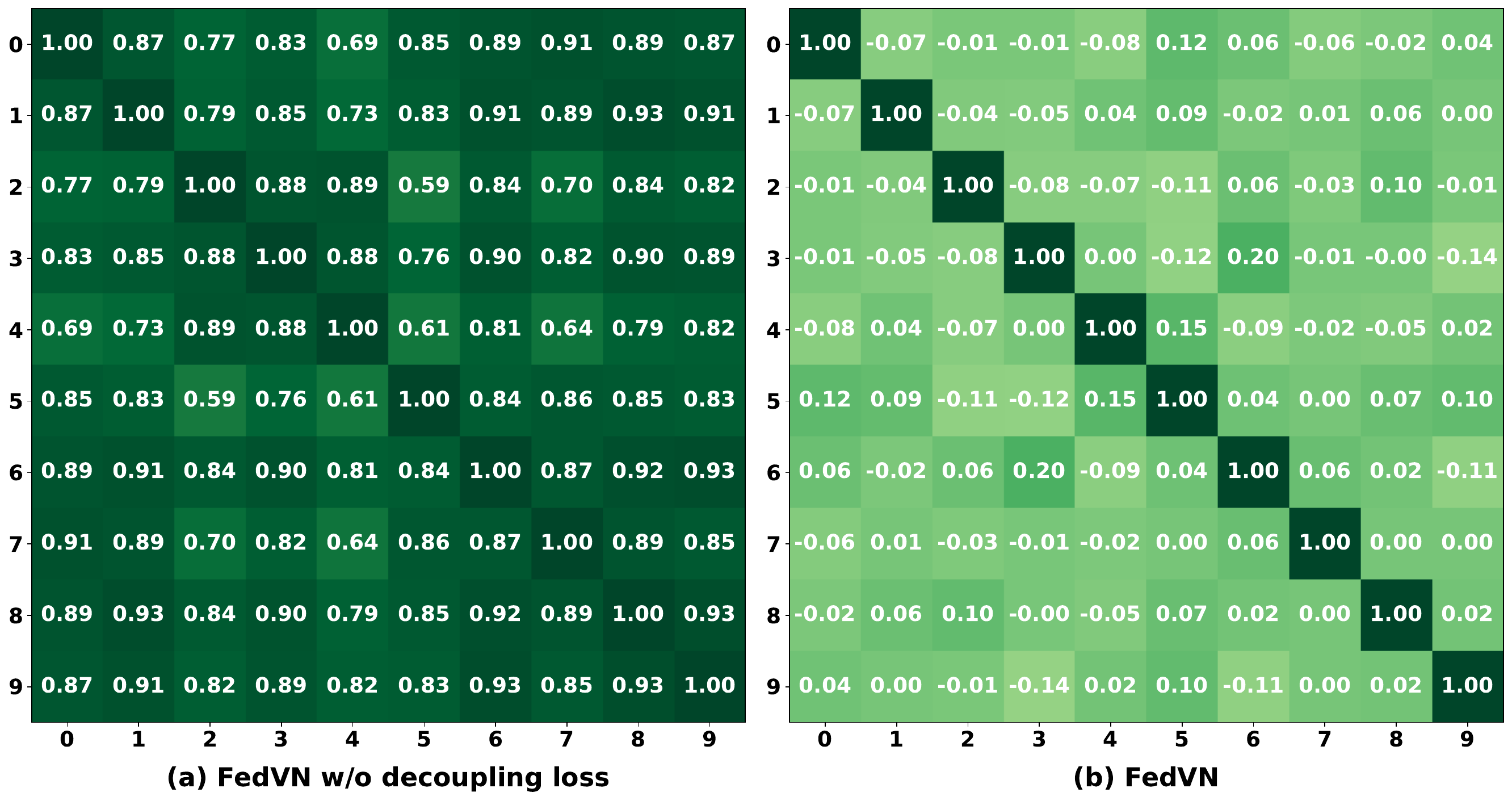}
\caption{Pairwise cosine similarities between 10 VNs on SST2/Length.}
\label{fig:similarity_sst2}
\end{figure}

\section{More experimental Results}
\subsection{Visualization of VN Collapse}
As discussed in Section 4.2, VNs are prone to collapse to fewer VNs, which hinders the effectiveness of multiple VNs in FedVN. In this experiment, we compare pairwise cosine similarities between VNs by FedVN and its variant without decoupling loss. Figure \ref{fig:similarity_zinc} and Figure \ref{fig:similarity_sst2} report the results of 10 VNs on Zinc/Scaffold and SST2/Length, respectively. We can observe that when FedVN learns VNs without decoupling loss, the pairwise cosine similarities are mostly close to 1, specifically on Zinc/Scaffold. With our proposed decoupling loss, FedVN can eliminate the correlation between VNs as indicated by the figures.

\begin{figure}[!t]
\centering
\includegraphics[width=0.85\linewidth]{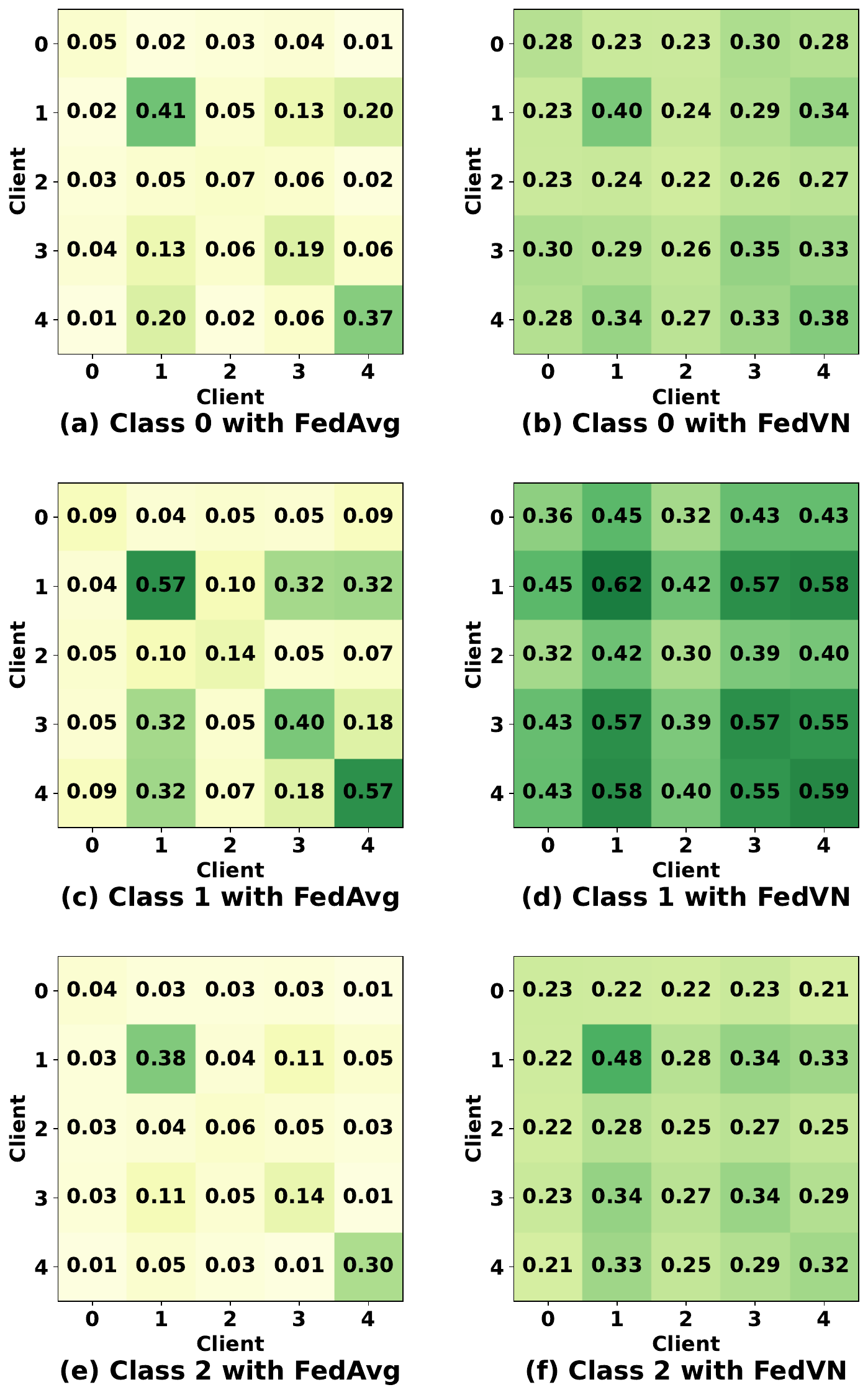}
\caption{Cross-client cosine similarities of graph embeddings in each client on Motif/Basis.}
\label{fig:figure_similarity_class}
\end{figure}

\subsection{Visualization of Distribution Shifts in FedAvg and FedVN}
FedVN aims to mitigate distribution shifts of graph data in FGL. We provide the visualization of cross-client cosine similarities of graph embeddings in each client on Motif/Basis in Figure \ref{fig:figure_similarity_class}. Motif/Basis is an artificial dataset where graphs are generated using five label-irrelevant base graphs (wheel, tree, ladder, star, and path) and three label-related motifs (house, cycle, and crane). Therefore, graphs are split into five clients (Client 0, 1, 2, 3, 4) by base graphs and are labeled as three classes (Class 0, 1, 2) by motifs. The visualization in Figure \ref{fig:figure_similarity_class} includes six heatmaps. Each entry in a heatmap represents the cross-client average cosine similarity between each pair of graph embeddings from the same class in the two clients. For instance, in heatmap (a), the value 0.20 represents the average cosine similarity between each pair of graph embeddings from Class 0 in Client 1 and Client 4. From the six heatmaps, we can observe that FedVN significantly improves cross-client cosine similarities of each class on Motif/Basis compared with FedAvg. Taking Client 4 as an example, graph embeddings by FedAvg in Client 4 are quite different from those in other clients, especially for Class 0 and 2. In contrast, FedVN can obtain more similar graph embeddings from the same class across clients. Therefore, we can conclude that FedVN effectively mitigates the distribution shift issue of graph data across clients in FGL.

\section{Limitations and future work} \label{limitation}
Although FedVN does not upload private data from clients to the server, we may still have privacy concerns about FedVN. First, FedVN shares VNs across clients. The VNs may contain private information, which might be attacked by others. Second, Malicious clients may perturb the feature matrix of VNs and consequently affect the performance of GNN models. Since this paper mainly focuses on model utility in FGL, we leave the exploration of privacy leakage as our future direction. In addition, FedVN introduces extra computational cost in FGL. One simple solution to this problem is reducing the size of edge generators. We will try to explore more efficient approaches to solve this problem.

\end{document}